%% file: sample-sigconf.tex
\documentclass[sigconf]{acmart}
\usepackage{booktabs}
\usepackage{amsmath}
\usepackage{mathrsfs}
\usepackage{subfigure}
\usepackage{pgfplots}\pgfplotsset{compat=1.9}
\usepackage{xcolor}
\usepackage{multirow}
\usepackage{bbm}
\usepackage{bm}
\usepackage{balance}
\usepackage[linesnumbered,boxed,ruled,vlined]{algorithm2e}

\copyrightyear{2021}
\acmYear{2021} 
\setcopyright{iw3c2w3}
\acmConference[WWW '21]{Proceedings of the Web Conference 2021}{April 19--23, 2021}{Ljubljana, Slovenia} 
\acmBooktitle{Proceedings of the Web Conference 2021 (WWW '21), April 19--23, 2021, Ljubljana, Slovenia}
\acmPrice{}
\acmDOI{10.1145/3442381.3449786}
\acmISBN{978-1-4503-8312-7/21/04}

\author[Y. Wang, Y. Min, X. Chen, J. Wu]{Yingheng Wang$^1$, Yaosen Min$^{2,*}$, Xin Chen$^{3,*}$, and Ji Wu$^{1,\dagger}$}

\makeatletter
\def\authornotetext#1{
	\g@addto@macro\@authornotes{%
	\stepcounter{footnote}\footnotetext{#1}}%
}
\makeatother

\authornotetext{The two authors contributed equally to this work.}
\authornotetext{To whom correspondence should be addressed.}

\affiliation{%
\institution{$^1$Department of Electronic Engineering, Tsinghua University}
\institution{$^2$Institute of Interdisciplinary Information Sciences, Tsinghua University}
\institution{$^3$Technology and Engineering Group, Tencent}
\institution{$^1${\em wangyh20@mails.tsinghua.edu.cn, wuji\_ee@mail.tsinghua.edu.cn}}
\institution{$^2${\em minys18@mails.tsinghua.edu.cn}}
\institution{$^3${\em marcuschen@tencent.com}}
}

\def\authors{Yingheng Wang, Yaosen Min, Xin Chen, Ji Wu}

\begin{document}
\title{Multi-view Graph Contrastive Representation Learning for Drug-Drug Interaction Prediction}

\begin{abstract}
Potential Drug-Drug Interactions (DDI) occur while treating complex or co-existing diseases with drug combinations, which may cause changes in drugs’ pharmacological activity. Therefore, DDI prediction has been an important task in the medical health machine learning community. Graph-based learning methods have recently aroused widespread interest and are proved to be a priority for this task. However, these methods are often limited to exploiting the inter-view drug molecular structure and ignoring the drug’s intra-view interaction relationship, vital to capturing the complex DDI patterns. This study presents a new method, multi-view graph contrastive representation learning for drug-drug interaction prediction, MIRACLE for brevity, to capture inter-view molecule structure and intra-view interactions between molecules simultaneously. MIRACLE treats a DDI network as a multi-view graph where each node in the interaction graph itself is a drug molecular graph instance. We use GCN to encode DDI relationships and a bond-aware attentive message propagating method to capture drug molecular structure information in the MIRACLE learning stage. Also, we propose a novel unsupervised contrastive learning component to balance and integrate the multi-view information. Comprehensive experiments on multiple real datasets show that MIRACLE outperforms the state-of-the-art DDI prediction models consistently.
\end{abstract}



\maketitle

\input{samplebody-conf}

\begin{acks}
This work is jointly supported by National Key Research and Development Program of China (No.2018YFC0116800) and Beijing Municipal Natural Science Foundation (No.L192026).
\end{acks}

\bibliographystyle{ACM-Reference-Format}
\balance
\bibliography{sample-bibliography}

\end{document}

%% file: samplebody-conf.tex
\section{Introduction}
Drugs may interact with each other when drug combinations occur, which will alter their effects. Such situations can increase the risk of patients’ death or drug withdrawal, particularly in the elderly, with a prevalence of 20-40\%\cite{palleria2013pharmacokinetic}. Recent studies estimate that 6.7\% of the US hospitalized patients have severe adverse drug reactions with a fatality rate of 0.32\%\cite{lazarou1998incidence}. However, polypharmacy is inevitable because the concurrent use of multiple medications is necessary for treating diseases that are often caused by complex biological processes that resist the activity of any single drug. Hence, there is a practical necessity to identify the interaction of drugs.

Meanwhile, detection of DDI remains a challenging task: traditional wet chemical experiments are expensive and cumbersome, and too small in scale, limiting the efficiency of DDI detection. Besides, there are tedious and time-consuming clinical tests and patient follow-up afterward. These problems make it urgent to develop a new, computationally assisted DDI prediction method.

The machine learning methods facilitate the computer-assisted DDI prediction. Prior traditional machine learning approaches focus mainly on inputting various fingerprints or similarity-based features such as \cite{vilar2014similarity}. Among them, in the early years, research works like \cite{vilar2012drug} are designed simple, which only consider single fingerprint. After 2014, plenty of research works based on multi-source heterogeneous fingerprints flourish. \citet{10.1371/journal.pone.0196865} establishes a logistic regression model with similarity-based features that are constructed on side effects and physiological effects as inputs. Using features including multi-dimensional molecular structures, individual drug side effects, and interaction profile fingerprints, \citet{ryu2018deep} applies various deep learning-based methods to integrate different types of inputs and obtain drug latent representations to detect DDI.

\begin{figure}[htbp]
\centering
\includegraphics[width=3in]{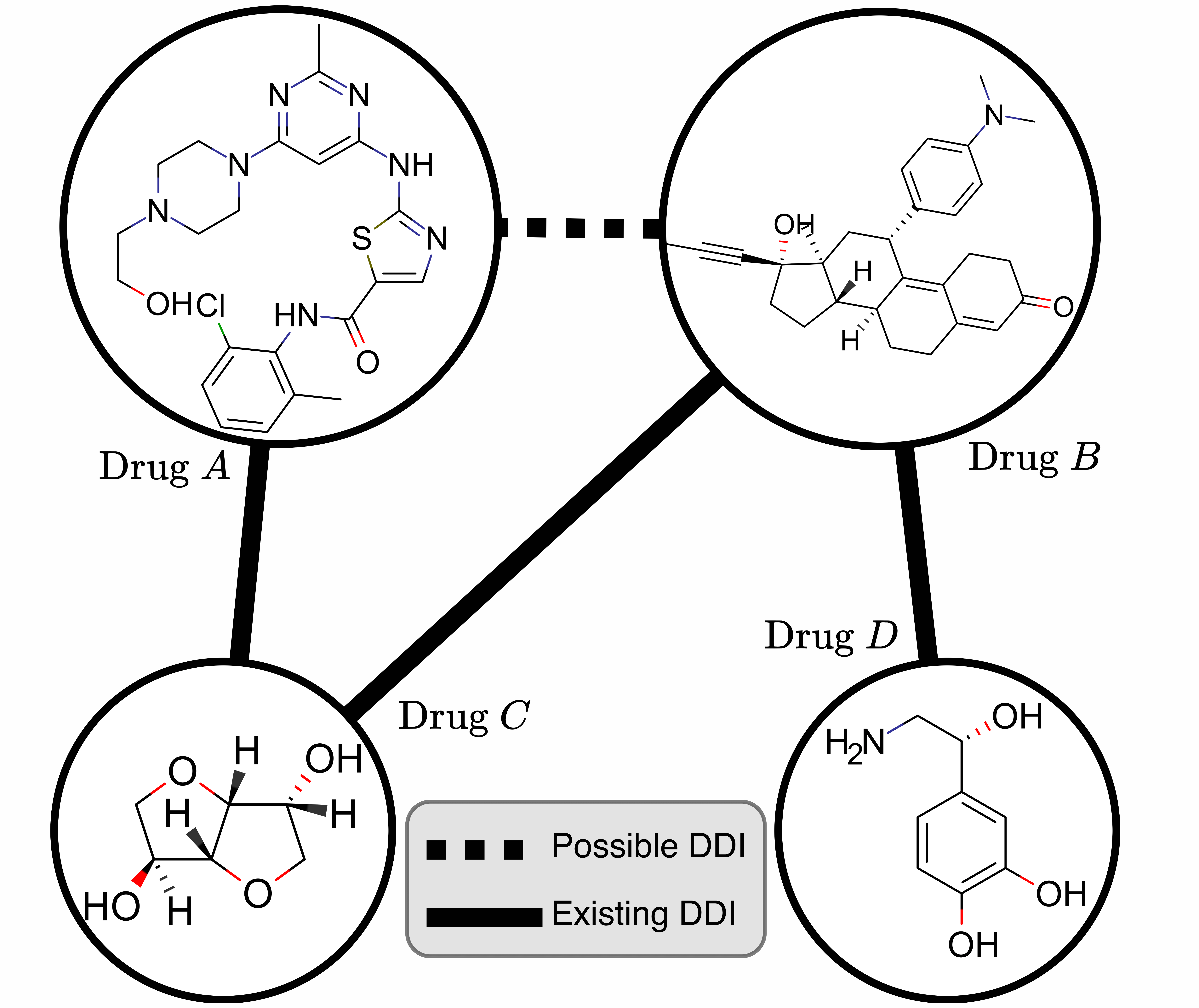}
\caption{A toy example of the multi-view graph.}
\label{fig:ddi}
\end{figure}

\begin{figure*}[htbp]
\centering
\includegraphics[width=\linewidth]{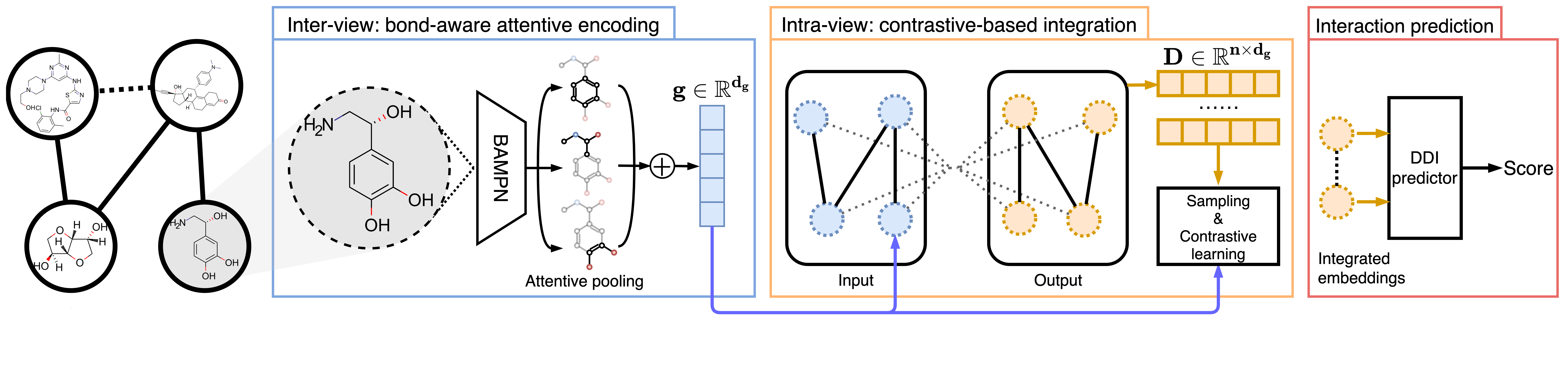}
\caption{The illustrative schematic diagram of our proposed framework MIRACLE. There are three sequential and interdependent phases: 1) In the inter-view, drug molecular graphs are encoded into drug embeddings by a bond-aware message passing network with attentive pooling. 2) In the intra-view, a GCN encoder is used to integrate the external DDI relationships into drug embeddings. Also, a contrastive learning-based method is applied to balance information from different views to update drug embeddings. 3) With the learned drug embeddings, an interaction predictor is designed to generate final prediction results.}
\label{fig:frame}
\end{figure*}

In recent years, many graph representation learning-based methods have been applied to extract features from raw molecular graph data. \citet{zitnik2018modeling} constructs a heterogeneous network including drug and protein entities and uses graph convolutional networks (GCN)\cite{kipf2016semi} to learn their latent representations and predict possible adverse drug side effects occurring in drug combinations. \citet{ma2018drug} proposes a method integrating multi-view similarity-based features by attention mechanism, where a view refers to a similarity matrix constructed by a feature and learns drug representations with graph autoencoder (GAE)\cite{kipf2016variational}. \citet{deac2019drug} applies a co-attention mechanism to compute the attentional coefficients among all atoms in a pair of drugs to learn drug-drug information jointly. \citet{chen2019drug} uses a siamese GCN to learn the pair-wise drug representations and make predictions with a similarity-based framework.  More recently, \citet{lin2020kgnn} merges several datasets into a vast knowledge graph with 1.2 billion triples. Through KGNN layers which embed 2-hop local structures of drugs, this model improves performance significantly. \citet{bai2020bi} proposes a bi-level graph attention network to encode drugs and make DDI predictions, which is similar as our method. However, it ignores the balance of information from different levels.

Although methods mentioned above achieve satisfactory results and become the state-of-the-art, they still have some limitations. \emph{Firstly}, previous models based on multiple similarity-based fingerprints require feature engineering or feature integration, which are time-consuming and challenging to collect\cite{xin2020research}. They mainly rely on the empirical assumption that chemically, biologically, or topologically similar drugs are more likely to interact with each other. However, plenty of interactive drug pairs may have relatively low pair-wise similarities, which can be misled by the model to wrong prediction results. Meanwhile, large-scale datasets may lack some features, which are crucial to DDI prediction but unavailable for most kinds of drugs. Therefore, these models \textbf{have a deficiency in the scalability and robustness}. \emph{Secondly}, in practice, the inter-view information included in the drug molecular graph and the intra-view DDI relationships are both important for the DDI prediction task. The inter-view information contains multiple atom and bond's features and structure information inside drug molecules. And the intra-view DDI relationships contain multiple hidden interaction patterns. However, previous graph-based works only concentrate on a single view of drugs\cite{xin2020research}. Thus, \textbf{the combination of multi-view information inside the DDI network is often overlooked} by these methods.

To overcome these limitations, we propose a novel method, \underline{M}ulti-v\underline{I}ew g\underline{RA}ph \underline{C}ontrastive representation \underline{LE}arning for drug-drug interaction prediction, MIRACLE for brevity, to explore rich multi-view information among drugs and generate accurate latent vectors of drugs. In such a multi-view graph setting, we treat each drug instance as a molecular graph and DDI relationships as an interaction graph, namely the inter- and intra-view, respectively. In the inter-view, we first encode drug molecular graphs into drug embeddings by bond-aware message passing networks with attentive pooling to fuse information from atom and bond features and molecule structures. Then, we integrate them with the intra-view information, the external DDI relationships to update drug embeddings. Also, for the multi-view information integrated into drug embeddings, it is necessary to balance information from different views. Thus, we use a contrastive learning-based component to tackle this problem. Finally, we obtain interaction embeddings through the learned
drug embeddings and make predictions. Figure \ref{fig:ddi} shows a multi-view graph in the context of a DDI network. Drug A, B, C, D denote four drugs in the DDI network. The solid and dashed lines indicate existing and possible interactions. The internal structure of each drug shows its molecular graph.

Figure \ref{fig:frame} illustrates the framework of the proposed MIRACLE method. At first, all drug instances are modeled as molecular graphs, where each node represents an atom, and each edge is a bond. Then, each molecular graph is proceeded successively. The inter-view latent vectors for all drugs involved in the DDI network can be obtained through bond-aware message passing neural networks with attentive pooling. After that, we integrate the inter-view and intra-view information to update drug representations by a GCN encoder where we treat drug instances as nodes, the inter-view drug embeddings as node features, and the external DDI relationships as graph connectivity. For balancing information from different views better, we design a novel contrastive learning component to optimize the model by setting a mutual information maximization objective. Finally, we use a two-layer fully-connected neural network to make predictions with the interaction embeddings calculated from the obtained drug embeddings. To help MIRACLE learn the commonality between two views well, we also make predictions using the inter-view drug representations and define a disagreement loss to enforce prediction results in two views to a consistency. The main contributions of this work are summarized as follows:
\begin{itemize}
    \item We model the DDI data in a multi-view graph setting and use graph representation learning methods to capture complex interaction patterns, which is a novel perspective and has rarely been studied before.
    \item To balance mult-view information in the drug representations, we do not rely on the supervised information only but design a novel contrastive learning component to update drug embeddings by a mutual information maximization objective.
    \item Extensive experiments conducted on a variety of real-world datasets show that MIRACLE evidently outperforms the state-of-art methods, even with few labeled training instances.
\end{itemize}

\section{Methodology}
This section introduces the proposed DDI prediction approach, which is an end-to-end representation learning model consisting of three sequential and interdependent phases. At first, we define a bond-aware message passing network with attentive pooling, which encodes drug molecular graph data to corresponding low-dimensional embedding vectors. The second is an information integration module that uses a GCN encoder to integrate the inter-view drug embedding vectors with the external DDI relationships. The last is an interaction predictor, which is responsible for predicting missing interactions in a DDI Network.

The proposed model takes the SMILES representations of corresponding drug molecules and the DDI network connectivity matrix as inputs. The SMILES representations will be converted into molecular graphs via RDKit\cite{landrum2013rdkit}. Meanwhile, we extract the atom list and multi-channel adjacency matrix from the molecular graphs, which will be fed into the following bond-aware message passing networks. In this section, we will depict the architecture, optimization objective, and training process.

\subsection{Notations and Problem Formulation}\label{notation}
Before presenting our proposed model, We summarize the important notations adopted in this paper. We use upper boldface letters for matrices (\emph{e.g.} $\textbf{A} \in {\mathbb{R}^{m \times n}}$), boldface letters for vectors (\emph{e.g.} $\textbf{h} \in {\mathbb{R}^d}$), normal characters for scalars (\emph{e.g.} $d_g$ for the dimension of molecule-level embedding, $d_h$ for the dimension of atom-level embedding), and calligraphic for sets (\emph{e.g.} $\mathcal{G}$).

Suppose that we have a graph $\mathcal{G}$ which is presented by $\mathcal{G} = \{\mathcal{V}, \mathcal{E}\}$ where $\mathcal{V}$ is the set of vertices and $\mathcal{E}$ is the set of edges. We denote the $i$th atom as $v_i \in \mathcal{V}$ and the chemical bond connecting the $i$th and $j$th atoms as $e_{ij} \in \mathcal{E}$.

\textbf{Problem statement.} The DDI prediction task can be defined as a link prediction problem on graph. Given $g$ drug molecular graphs $\mathcal{G} := \{G^{(i)}\}^g_{i=1}$ and DDI network $\mathcal{N} = (\mathcal{G}, \mathcal{L})$ where $\mathcal{L}$ denotes the interaction links, the task link prediction can be defined as, for the network $\mathcal{N}$, predicting the existence of missing links.

\subsection{Bond-aware Message Passing Networks with Attentive Pooling}\label{MPN}
Since the molecular graphs are complex irregular-structured data to handle, in this subsection, we use a bond-aware message passing network(BAMPN)\cite{gilmer2017neural} to map nodes to real-value embedding vectors in the low-dimensional space. We equipped our model with propagation-based message passing layers according to simple chemistry knowledge and a graph readout layer with attentive pooling\cite{vaswani2017attention} to generate a graph-level representation.

A molecule can be served as a graph where each atom is represented as a node and each bond as an edge. We construct the node information matrix by stacking randomly initialized embedding vectors for each atom considering its nuclear charge number and a multi-channel adjacency matrix whose channel dimension indicates different chemical bond types, including single, double, triple and aromatic bond.

Given the node information matrix and multi-channel adjacency matrix as inputs, we encode them by two successive processes of message passing. The first phase can be described using the following message function:

\begin{equation}
\bf{\tilde{h}}_i^{(l)} = \sum_{j \in {\mathcal{C}(i)}} \bf{W}_{c_{ij}}^{(l)}\bf{h}_j^{(l-1)}\label{message}
\end{equation}
where $\bf{W}_{c_{ij}}^{(l)} \in \mathbb{R}^{d_h \times d_h}$ is a matrix of trainable parameters shared by the same type of chemical bond $c_{ij}$ at the $l$th layer, $\bf{\tilde{h}}_i^{(l)}$ represents the candidate hidden state at the $l$th layer for node $v_i$, $\bf{h}_j^{(l-1)} \in \mathbb{R}^{d_h}$ represents the hidden state at the $(l-1)$th layer for the neighboring node $v_j$, and $\mathcal{C}(i)$ denotes the neighboring nodes of the center node $v_i$.

The equation (\ref{message}) shows that the node information corresponding to the same type of chemical bond share parameters during the affine transformation. The chemical interpretability for this massage function is straightforward: neighboring nodes connected by the same type of chemical bond have similar effects, and vice versa.

Given $\bf{\hat{h}}_i^{(l)}=[\bf{h}_i^{(l-1)};\bf{\tilde{h}}_i^{(l)}]$ where $[;]$ denotes the concatenation operation, for the second phase, inspired by \cite{srivastava2015highway}, we additionally define three non-linear transforms $\mathit{F}(\bf{W}_f, \bf{\hat{h}}_i^{(l)})$, $\mathit{T}(\bf{W}_t, \bf{\hat{h}}_i^{(l)})$, and $\mathit{C}(\bf{W}_c, \bf{\hat{h}}_i^{(l)})$. Thus, this process can be described as the following update function:
\begin{equation}
\begin{aligned}
\bf{h}_i^{(l)} = \mathit{T}(\bf{W}_t, \bf{\hat{h}}_i^{(l)}) \odot \mathit{F}(\bf{W}_f, \bf{\hat{h}}_i^{(l)}) + \mathit{C}(\bf{W}_c, \bf{\hat{h}}_i^{(l)}) \odot \bf{h}_i^{(l-1)}\label{}
\end{aligned}
\end{equation}
where $\bf{h}_i^{(l)}$ represents the hidden state at the $l$th layer for node $v_i$, and $\odot$ denotes the element-wise product. We refer to $\mathit{F}$ as the \emph{fuse} gate, $\mathit{T}$ as the \emph{transform} gate, and $\mathit{C}$ as the \emph{carry} gate, since they express how much of the hidden state is produced by transforming the fusion of the candidate and previous hidden state and carrying it, respectively. We can take both the influence on the concentrated center node exerted by neighboring ones and itself at the previous layer into consideration through this update process.

We stack several message passing layers ($L$ in total) to learn the hidden representation for every node/atom $v$ in a molecular graph $\mathcal{G}$ and obtain the final hidden states for each atom at the last message passing layer. To make predictions with drug representations, we need to generate an embedding vector for each molecular graph. Therefore, we apply a simple but efficient attentive readout layer inspired by \citet{li2015gated} as follows:

\begin{equation}
    \bf{a}_i = tanh(\bf{W}_a[\bf{h}_i^{(0)};\bf{h}_i^{(L)}] + \bf{b}_a)
\end{equation}
\begin{equation}\label{eq:inter}
    \bf{g} = \sum_{v_i \in \mathcal{V}}
    \bf{a}_i \odot (\bf{W}_o\bf{h}_i^{(L)} + \bf{b}_o)
\end{equation}
where $[;]$ denotes the concatenation operation, $tanh(\cdot)$ denotes the tanh activation function, attention score $\bf{a}_i \in \mathbb{R}^{d_g}$ denotes the importance score of the atom $v_i$, $\odot$ denotes the element-wise product, and $\bf{g} \in \mathbb{R}^{d_g}$ is the obtained embedding vector for the molecular graph. Stacking the embedding vectors of drugs over given dataset, we get the inter-view embedding matrix $\bf{G} \in \mathbb{R}^{n \times d_g}$. It should be noted that all of the parameters are shared across all the atoms.

\subsection{GCN for Integrating Multi-view Network Information}
For drug molecular graphs, after the non-linear mapping introduced above, we get low-dimensional representations of them. In order to encode the intra-view interaction information into the obtained inter-view drug embeddings, we establish a graph-based encoder to integrate the multi-view network information. Recently many graph representation learning-based methods such as \citet{kipf2016semi} have demonstrated their superiority to traditional graph-based methods such as various dimensionality reduction algorithms\cite{yan2006graph}. Therefore, in this paper, we make use of GCN in terms of efficiency and effectiveness. In the following, we utilized a multi-layer GCN($M$ in total) encoder to smooth each node's features over the graph's topology. In this context, we refer to the node's features as low-dimensional representations of drugs obtained from the previous bond-aware message passing networks and the graph's topology as the interaction relationship inside the DDI network.

Assumed the number of drugs in the DDI network is denoted by $n$, formally, we are given the adjacency matrix of DDI network $\bf{A} \in \mathbb{R}^{n \times n}$ and the attribute matrix $\bf{G} \in \mathbb{R}^{n \times d_g}$ of the DDI network $\mathcal{N}$ as inputs. Before the graph convolution operation, we normalized the adjacency matrix $\bf{A}$:
\begin{equation}
    \bf{\hat{A}} = \bf{\tilde{K}}^{-\frac{1}{2}}(\bf{A}+\bf{I}_n)\bf{\tilde{K}}^{-\frac{1}{2}}
\end{equation}
where $\bf{I}_n$ represents the identity matrix and $\bf{\tilde{K}}_{ii} = \sum_j(\bf{A}+\bf{I}_n)_{ij}$. Then we apply the GCN encoder framework as follows:
\begin{equation}\label{eq:intra}
\begin{aligned}
    \bf{D}^{(1)} = \mathit{U}(\bf{A}, \bf{G}, \bf{W}_u^{(0)}, \bf{W}_u^{(1)})
    = \bf{\hat{A}}ReLU(\bf{\hat{A}}\bf{G}\bf{W}_u^{(0)})\bf{W}_u^{(1)}
\end{aligned}
\end{equation}
where $\bf{W}_u^{(0)} \in \mathbb{R}^{d_u^{(0)} \times d_u^{(1)}}$ and $\bf{W}_u^{(1)} \in \mathbb{R}^{d_u^{(1)} \times d_u^{(2)}}$ are two learnable weight parameters at the $0^{th}$ and $1^{th}$ layer of the GCN encoder, respectively. The second dimension of the learnable weight parameter at the last layer of the GCN encoder is set to $d_g$. Through the GCN encoder, then we get the intra-view embedding matrix $\bf{D} \in \mathbb{R}^{n \times d_g}$ for drugs in the DDI network $\mathcal{N}$.

\begin{figure}[htbp]
\centering
\includegraphics[width=3.2in]{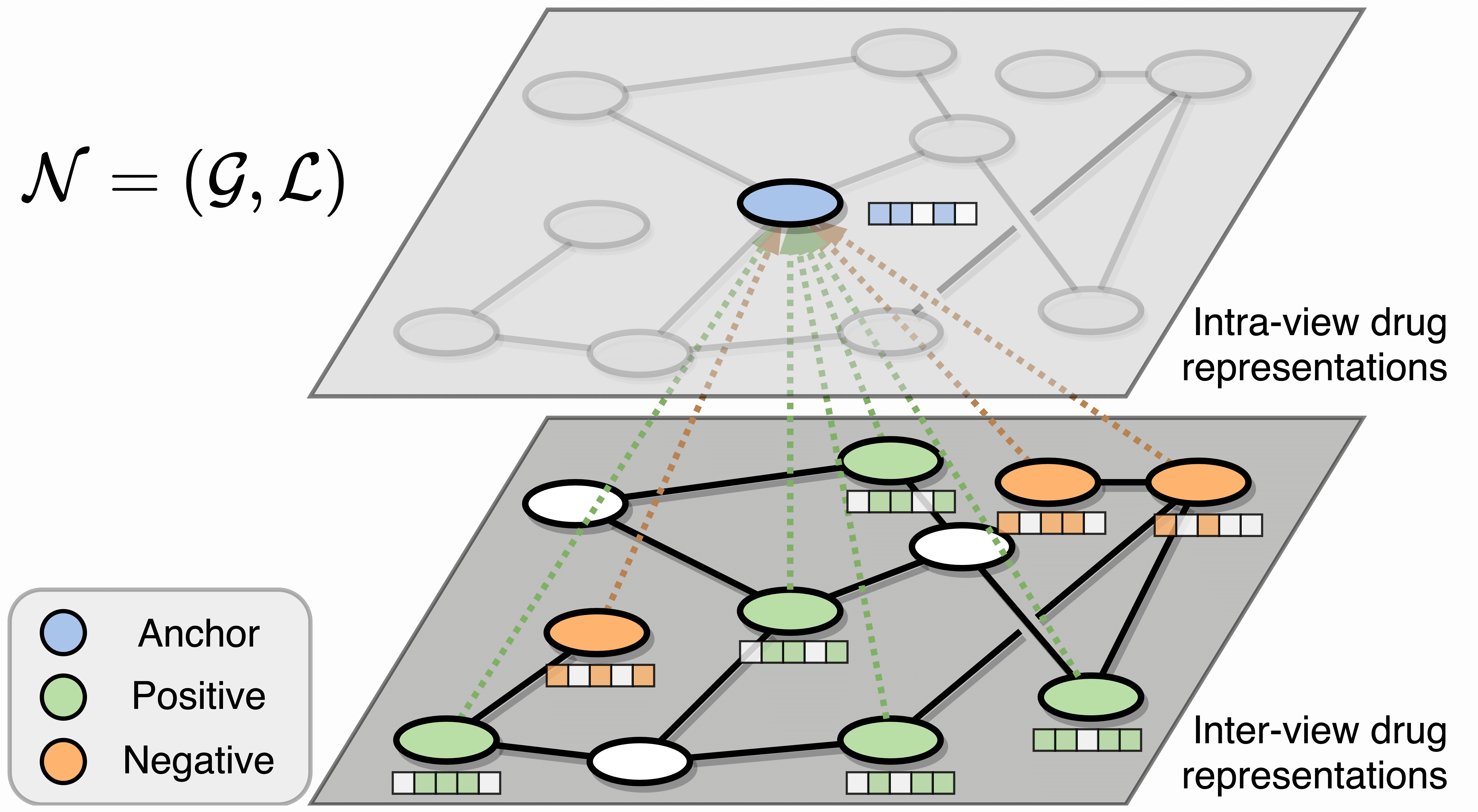}
\caption{The proposed graph contrastive learning framework.}
\label{fig:contrastive}
\end{figure}

\subsection{Contrastive Learning of Drug Representation}\label{CLDR}
As stated in \citet{li2018deeper}, the graph convolution operation can be considered Laplacian smoothing for nodes' features over graph topology. The Laplacian smoothing computes the new node features as the weighted average of itself and its neighboring nodes. On the one hand, although it helps make nodes in the same cluster tend to have similar representations, it may cause the over-smoothing problem and make nodes indistinguishable. On the other hand, it will concentrates so much on node features information that makes the obtained embeddings lack structural information.

Furthermore, the inter-view drug embeddings are learned directly from molecular graphs, including raw attributes and structural information. Specifically, the representations may contain multiple functional groups' information inside a molecule, different local connectivity constructed by various combinations of the same atoms and bonds, etc. Such information has a significant influence on interaction predictions. However, after graph convolutions, the inter-view features are smoothed over the whole DDI network topology and become blurred. Therefore, it is necessary to balance multi-view information in the drug embeddings. In this subsection, we propose a novel graph contrastive learning framework to handle these issues.

Figure \ref{fig:contrastive} illustrates the proposed graph contrastive learning component. In MIRACLE, we naturally have two different graph views in the multi-view graph setting to learn drug representations by maximizing the agreement between inter-view and intra-view embeddings. We define our mutual information(MI) estimator on inter- and intra-view pairs, maximizing the estimated MI over the given dataset $\mathcal{N}$. To be specific, for each drug $i$, fixing itself as an anchor $A$, we get a set of positive samples $\mathbb{P}$ which is made up of itself and its $k$-order neighboring nodes, and a set of negative samples $\mathbb{N}$ from nodes not in its $k$-hop neighbors. Considering the assumption that representations of drugs that interact with the same drug may contain similar information, we usually set $k=1$. Then, we generate each positive pair $(A, P)$ where $P \in \mathbb{P}$ and negative pair $(A, N)$ where $N \in \mathbb{N}$. After iterations, we can obtain all possible positive pairs and negative pairs. Afterward, we employ a contrastive objective that enforces the intra-view representation of the anchor $A$ agree with the inter-view representations of the positive samples and can be distinguished from the inter-view representations of the negative samples. The contrastive objective is formulated as follows:
\begin{equation}\label{eq:con}
\begin{aligned}
    \hat{\omega}, \hat{\phi}, \hat{\psi} =  \mathop{argmax}\limits_{\omega, \phi, \psi}\sum\limits_{i \in \mathcal{G}}\frac{1}{\tilde{\mathcal{C}}(i)}\sum\limits_{j \in \mathcal{C}(i) \cap \{i\}}\mathcal{\hat{I}}_{\omega, \phi, \psi}(\bf{g}_j^\phi; \bf{d}_i^\psi)
\end{aligned}
\end{equation}
where $\tilde{C}(i) := | \mathcal{C}(i)+1 |$, $\phi$ and $\psi$ denote the set of parameters of the BAMPN and GCN encoder, respectively, $\mathcal{\hat{I}}_{\omega, \phi, \psi}$ is the mutual information estimator modeled by discriminator $\mathcal{T}_{\omega}$ and parameterized by a neural network with parameters $\omega$. As stated in \citet{hjelm2018learning}, contrastive learning-based methods concentrate primarily on the maximization of mutual information(MI) instead of its precise value. Therefore, following \citet{nowozin2016f}, we use a lower-bound to the MI on the Jensen-Shannon-based representation of the KL-divergence for better performance in practice, which can be formulated as:
\begin{equation}
\begin{aligned}
    \mathcal{\hat{I}}_{\omega, \phi, \psi}(\bf{g}_j^\phi; \bf{d}_i^\psi) :&=  \mathbb{E}_{\mathbb{D}}[-sp(-\mathcal{T}_{\omega, \phi, \psi}(\bf{g}_j^\phi(x); \bf{d}_i^\psi(x)))] \\
    & - \mathbb{E}_{\mathbb{D} \times \tilde{\mathbb{D}}}[sp(\mathcal{T}_{\omega, \phi, \psi}(\bf{g}_j^\phi(x'); \bf{d}_i^\psi(x)))]
\end{aligned}
\end{equation}
where $x$ is an input sample, $x'$ is an input sample from $\tilde{\mathbb{D}} = \mathbb{D}$, $\mathbb{D}$ denotes an empirical probability distribution, and $sp(\cdot)$ is the softplus function. It should be noted that we generate negative samples using all possible combinations of inter/intra-view embeddings across all drugs in a batch, which means that negative pairs are sampled across all drugs. Since $\bf{d}_i^\psi$ is encouraged to have high MI with samples containing information at both views, this favors encoding aspects of the data shared across different samples and views.

\subsection{Drug-drug Interaction Prediction}
\subsubsection{Interaction predictor}
For each interaction link $l_{ij} \in \mathcal{L}$, We first compress two drug embedding vectors into an interaction link embedding vector:
\begin{equation}
    \bf{l} = \bf{d}_i \odot \bf{d}_j
\end{equation}
where $\odot$ denotes the element-wise product, $\bf{l}$ is the interaction link embedding vector. Then, we apply a two-layer fully-connected neural network to make the final prediction:
\begin{equation}
    \bf{p} = \sigma(\bf{W}_p\mathop{ReLU}(\bf{W}_l\bf{l} + \bf{b}_l) + \bf{b}_p)
\end{equation}
where $\bf{p} \in \mathbb{R}^k$ and $[;]$ denotes the concatenation operation. If the aim is to predict the occurrence of DDI, $k$ is 2.

Besides, we design another auxiliary interaction predictor using the inter-view drug embeddings. We cascade the last layer of the MPN with a fully connected layer and a sigmoid transformation function to construct this classifier. The prediction of the inter-view interaction predictor is denoted as  $\bf{r} \in \mathbb{R}^k$. Optimizing the inter-view interaction prediction results can help the supervised information directly flow into previous network layers. The model learns the commonality between different views through a disagreement loss, which will be discussed below. So, finally, we get two prediction results corresponding to two different predictors. It should be noted that we only use $\bf{p}$ for the final DDI prediction.

\subsubsection{Training}
Both predictors' primary goal is to minimize the supervised loss, which measures the distance between the predictions and the true labels. Another goal is to minimize a disagreement loss, which measures the distance between two predictors' predictions. The purpose of minimizing this disagreement loss is to enforce the model to pay more attention to the commonality between two different views and consistency between two predictors.

Formally, we formulate the supervised loss for the labeled interaction links and the disagreement loss for the unlabeled interaction links:
\begin{equation}
    \mathscr{L}_s = \sum\limits_{l_i \in \mathcal{L}_l}(\mathscr{C}(\bf{r}_i, y_i)+\mathscr{C}(\bf{p}_i, y_i))
\end{equation}
\begin{equation}
    \mathscr{L}_d = \sum\limits_{l_j \in \mathcal{L}_u}\mathscr{K}(\bf{p}_j || \bf{r}_j)
\end{equation}
where $y_i$ is the true label of $l_i$, $\mathcal{L}_l$ and $\mathcal{L}_u$ denote the labeled and unlabeled links in $\mathcal{L}$ respectively, $\mathscr{C}(\cdot, \cdot)$ is the cross-entropy loss function and $\mathscr{K}(\cdot || \cdot)$ is the Kullback-Leibler divergence.

With contrastive loss $\mathscr{L}_c$, supervised loss $\mathscr{L}_s$ and disagreement loss $\mathscr{L}_d$, the objective function of our model is,
\begin{equation}\label{loss}
    \mathscr{L} = \mathscr{L}_s + \alpha \mathscr{L}_c + \beta \mathscr{L}_d
\end{equation}
where $\alpha$ and $\beta$ are hyper-parameters for the trade-off for different loss components. With the objective function, we use the back-propagation algorithm to find the best solution for the trainable parameters $\omega, \phi, \psi$.

\subsubsection{Computation}
The key parts of our model include the GNNs-based pipeline and the contrastive learning component.
Regarding GNNs, the cores are to compute the inter-view molecular graph embeddings matrix $\bm{G}$ after equation (\ref{eq:inter}) and the drug embeddings matrix $\bm{D}$ in equation (\ref{eq:intra}). The computations require $O(NCu+Ld)$ where $N$ denotes the number of drug instances, $C$ denotes the number of chemical bonds in the drug instance, $L$ is the number of DDI, $u$ indicates the input feature dimension, and $d$ is the feature dimension of the drug embeddings.
Next, the computation by the contrastive learning component in equation (\ref{eq:con}) requires $O(2kN)$ flops in total, where $k$ denotes a fixed number of positive samples.
Overall, the complexity of our method is $O(N(Cu+2k)+Ld)$, which scales linearly in terms of the number of drug instances(i.e., $N$), the number of DDI(i.e., $L$), and the number of chemical bond in each drug instance(i.e., $C$).

In molecular graphs, the number of chemical bonds is usually under 30. Thus, the complexity mostly depends on $N$ and $L$ when extending this method to handle DDI networks with millions of drugs.
We apply full-batch gradient descent in this study because known DDI are limited, which means the sizes of DDI datasets are far under thousands of drugs and millions of interactions. So scalability is not an essential issue in our task. If our method is extended to other tasks, as seen, our model is computationally comparable to the GCN-based baselines\cite{wu2020comprehensive}. 

\section{Experiments}
In this section, we first describe the datasets, compared methods, and evaluation metrics used in the experiments. Then, we compare the proposed MIRACLE with other comparative methods. Finally, we make detailed analysis of MIRACLE under different experimental settings.

\subsection{Datasets}
We evaluate the proposed method on three benchmark datasets, i.e., \emph{ZhangDDI}\footnote{https://github.com/zw9977129/drug-drug-interaction/tree/master/dataset}, \emph{ChCh-Miner}\footnote{http://snap.stanford.edu/biodata/datasets/10001/10001-ChCh-Miner.html} and \emph{DeepDDI}\footnote{https://zenodo.org/record/1205795} with different scales for verifying the scalability and robustness of our model. These three datasets are small-scale, medium-scale, and large-scale, respectively. The \emph{ZhangDDI} dataset contains a relatively small number of drugs where all the fingerprints are available for all drugs. However, for \emph{DeepDDI}, the large-scale one, many fingerprints are missing in most drugs. For the \emph{ChCh-Miner} dataset, although it has almost three times the number of drugs in the \emph{ZhangDDI} dataset, it only has the same number of labeled DDI links. In our preprocessing, we remove the data items that cannot be converted into graphs from SMILES strings. The statistics of datasets are summarized as follows:
\begin{itemize}
    \item \emph{ZhangDDI}\cite{zhang2017predicting}: This dataset contains 548 drugs and 48,548 pairwise DDI and multiple types of similarity information about these drug pairs. 
    \item \emph{ChCh-Miner}\cite{biosnapnets}: This dataset contains 1,514 drugs and 48,514 DDI links without similarity-based fingerprints and polypharmacy side-effect information of each drug pair.
    \item \emph{DeepDDI}\cite{ryu2018deep}: This dataset contains 192,284 pair-wise DDI and their polypharmacy side-effect information extracted from DrugBank\cite{wishart2018drugbank}.
\end{itemize}
It should be noted that we conduct data removal because of some improper drug SMILES strings in Drugbank, which can not be converted into molecular graphs. The errors include so-old storage format of SMILES strings, wrong characters, etc.

\subsection{Comparing Methods}
To demonstrate the superiority of our proposed model, we implement many baseline approaches to compare their performance. The compared baselines cover similarity-based methods and graph-based methods. For the latter, to compare methods using different views of information reasonably, we define them under the same architecture as MIRACLE, which is summarized in Table \ref{tabmethod} and detailed as follows:
\begin{itemize}
    \item \textbf{Nearest Neighbor}\cite{vilar2012drug}: Vilar and his team used known interactions between drugs and similarity derived from substructure to conduct DDI prediction. We refer to the model as \textbf{NN} for simplicity.
    \item \textbf{Label Propagation}: \citet{zhang2015label} utilized the label propagation(LP) algorithm to build three similarity-based predictive models. The similarity is calculated based on the substructure, side effect, and off-label side effect, respectively. We refer to the three models as \textbf{LP-Sub}, \textbf{LP-SE} and \textbf{LP-OSE}, respectively.
    \item \textbf{Multi-Feature Ensemble}: \citet{zhang2017predicting} employed neighbor recommendation(NR) algorithm, label propagation(LP) algorithm, and matrix disturbs (MD) algorithm to build a hybrid ensemble model. The ensemble model exploited different aspects of drugs. We name the model as \textbf{Ens}.
    \item \textbf{SSP-MLP}: \citet{ryu2018deep} applied the combination of precomputed low dimensional Structural Similarity Profile(SSP) and Multi-layer Perceptron to conduct the classification. We refer to the model as \textbf{SSP-MLP}.
    \item \textbf{GCN}: \citet{kipf2016semi} used a graph convolutional network(GCN) for semi-supervised node classification tasks. We apply \textbf{GCN} to encode drug molecular graphs and use their representations to make predictions as a baseline.
    \item \textbf{GIN}: \citet{xu2018powerful} proposed a graph isomorphism network (GIN) to learn molecules' representations in various single-body property prediction tasks. We use \textbf{GIN} to encode drug molecular graphs and use their representations to make predictions as a baseline.
    \item \textbf{Attentive Graph Autoencoder}: \citet{ma2018drug} designed an attentive mechanism to integrate multiple drug similarity views, which will be fed into a graph autoencoder to learn the embedding vector for each drug. We refer to the model as \textbf{AttGA} and make predictions based on the learned drug representations pairwise as a baseline.
    \item \textbf{GAT}: \citet{velivckovic2017graph} utilized a graph attention network(GAT) to learn node embeddings by a well-designed attention mechanism on the graph. We use \textbf{GAT} to obtain drug embeddings based on the DDI network for predictions.
    \item \textbf{SEAL-CI}: \citet{li2019semi} firstly applied a hierarchical graph representation learning framework in semi-supervised graph classification tasks. We name this model as \textbf{SEAL-CI} and use the model to learn drug representations for DDI predictions as a baseline.
    \item \textbf{NFP-GCN}: \citet{duvenaud2015convolutional} is the first graph convolution operator, which is specific to molecules. We named the model as \textbf{NFP-GCN}. We change our bond-aware message passing networks into NFP to be a baseline.
\end{itemize}
\begin{table*}[t]
\centering
\caption{Comparison of baseline methods.}
\begin{tabular}{c|cccc}
	\toprule
    \textbf{Algorithm} & \textbf{Model Type} & \textbf{The Inter-view Model} & \textbf{The Intra-view Model} & \textbf{Feature Type} \\
    \hline
    \textbf{NN} & similarity-based & N/A & N/A & similarity-based fingerprint \\
    \textbf{LP} & similarity-based & N/A & N/A & similarity-based fingerprint \\
    \textbf{Ens} & similarity-based & N/A & N/A & similarity-based fingerprint \\
    \textbf{SSP-MLP} & similarity-based & N/A & N/A & similarity-based fingerprint \\
    \textbf{GCN} & inter-view & GCN & N/A & Molecular Graph \\
    \textbf{GIN} & inter-view & GIN & N/A & Molecular Graph \\
    \textbf{AttGA} & intra-view & N/A & AttGA & Interaction Relationship \\
    \textbf{GAT} & intra-view & N/A & GAT & Interaction Relationship \\
    \textbf{SEAL-CI} & multi-view & GCN & GCN & Molecular Graph \& Interaction Relationship \\
    \textbf{NFP-GCN} & multi-view & NFP & GCN & Molecular Graph \& Interaction Relationship \\
    \textbf{MIRACLE} & multi-view & BAMPN & GCN & Molecular Graph \& Interaction Relationship \\
	\bottomrule
\end{tabular}
\label{tabmethod}
\end{table*}

\subsection{Evaluation Metrics and Experimental Settings}
We divide the entire interaction samples into a train set and a test set with a ratio about $4:1$, and randomly select $1/4$ of the training dataset as a validation dataset. Note that we have only reliable positive drug pairs in the dataset, which means that only DDI certainly occur are recorded. We regard the same number of sampled negative drug pairs as the negative training samples for simplicity\cite{chen2019drug}.

We set each parameter group's learning rate using an exponentially decaying schedule with the initial learning rate $0.0001$ and multiplicative factor $0.96$. For the proposed model's hyper-parameters, we set the dimension of the hidden state of atoms and drugs as 256. The total number of the bond-aware message passing neural networks and the GCN encoder is 3. The coefficients $\alpha$ and $\beta$ in objective functions are set to 100 and 0.8, respectively, making the model achieve the best performance. To further regularise the model, dropout\cite{srivastava2014dropout} with $p = 0.3$ is applied to every intermediate layer's output.

We implement our proposed model with Pytorch 1.4.0\cite{paszke2019pytorch} and Pytorch-geometric 1.4.2\cite{fey2019fast}. Models are trained using Adam\cite{kingma2014adam} optimizer. The model is initialized using Xavier\cite{glorot2010understanding} initialization. We choose three metrics to evaluate our proposed model's effectiveness: \emph{area under the ROC curve(AUROC), area under the PRC curve(AUPRC)}, and \emph{F1}. We report the mean and standard deviation of these metrics over ten repetitions.

\subsection{Experimental Results}
We conduct experiments on three datasets with different characteristics to verify our proposed method's effectiveness in different scenarios. The three parts of the experiments validate the superiority of our MIRACLE method compared to baselines on the small scale dataset with all types of drug features and the medium scale dataset with few labeled DDI links, and the large scale dataset with missing drug features, respectively.

\subsubsection{Comparison on the \emph{ZhangDDI} dataset}
Table \ref{tabsmall} compares our MIRACLE model's performance against baseline approaches on the \emph{ZhangDDI} dataset, where almost all types of drug features can be used for the DDI prediction task. The best results are highlighted in boldface. MIRACLE integrates multi-view information into drug representations. In this model, we jointly consider the inter-view drug molecular graphs and the intra-view DDI relationships. According to the experiments, the proposed model achieves more excellent performance compared to these baseline approaches.

The performance of algorithms utilizing similarity-based fingerprints like \textbf{NN}, \textbf{LP}, and \textbf{SSP-MLP} is relatively poor, which only incorporate one kind of very important features. However, contrary to them, \textbf{Ens} obtains better results because of the combination of three distinct models utilizing eight types of drug feature similarities coupling with another six types of topological ones, demonstrating the importance of integrating information from multiple sources like similarity-based fingerprints and topological features.

Some graph-based methods perform worse than the models mentioned above because they only rely on the single view graph information. \textbf{GCN} and \textbf{GIN} encode drug molecular graphs by two different graph neural network frameworks. They make DDI predictions pairwise based on the obtained drug molecule representations.  \textbf{AttGA} and \textbf{GAT} directly learn drug representations from DDI interaction relationships and make predictions using the inner product results of target drug pairs' embeddings. The former acquires multiple connectivities of the DDI network according to different similarity matrices and applies a GCN encoder to obtain drug representations with varying relationships of interaction. This method fuses these drug representations to make final predictions that distinctly can achieve better performance than \textbf{GAT}, who only considers the DDI network's link existence.

The compared baselines in the multi-view graph settings like \textbf{SEAL-CI} and \textbf{NFP-GCN} outperform other baselines, demonstrating the integration of multi-view graph can improve the performance of models significantly. However, their performance is still inferior to that of the proposed method. Compared with the state-of-art method, MIRACLE further considers the importance of the message passing mechanism in terms of chemical bonds inside drug molecular graphs and the balance between multi-view graph information, which can learn more comprehensive drug representations. Whereas in \textbf{SEAL-CI} and \textbf{NFP-GCN}, they explicitly model the multi-view graphs and obtain the drug representations through a continuous graph neural network pipeline, with information equilibrium between different views ignored. Besides, \textbf{MIRACLE} adopts the self-attentive mechanism to generate an inter-view drug representation that automatically selects the most significant atoms that form meaningful functional groups in DDI reactions and neglect noisy, meaningless substructures.

\begin{table}
\centering
\caption{Comparative evaluation results on \emph{ZhangDDI}}
\begin{tabular}{c|ccc}
	\toprule
	\multirow{2}[0]{*}{\textbf{Algorithm}} & \multicolumn{3}{c}{\textbf{Performance}} \\
\cline{2-4}
		& \textbf{AUROC} & \textbf{AUPRC} & \textbf{F1} \\ \hline
	NN & \(67.81 \pm 0.25 \)& \(52.61 \pm 0.27\)& \(49.84 \pm 0.43\)  \\
	LP-Sub & \(93.39 \pm 0.13 \)& \(89.15 \pm 0.13\)& \(79.61 \pm 0.16\)  \\
	LP-SE & \(93.48 \pm 0.25 \)& \(89.61 \pm 0.19\)& \(79.83 \pm 0.61\)  \\
	LP-OSE & \(93.50 \pm 0.24 \)& \(90.31 \pm 0.82\)& \(80.41 \pm 0.51\)  \\
	Ens & \(95.20 \pm 0.14 \)& \(92.51 \pm 0.15\)& \(85.41 \pm 0.16\)  \\
	SSP-MLP & \(92.51 \pm 0.15 \)& \(88.51 \pm 0.66\)& \(80.69 \pm 0.81\)  \\
	GCN & \(91.91 \pm 0.62 \)& \(88.73 \pm 0.84\)& \(81.61 \pm 0.39\)  \\
	GIN & \(81.45 \pm 0.26 \)& \(77.16 \pm 0.16\)& \(64.15 \pm 0.16\)  \\
	AttGA & \(92.84 \pm 0.61 \)& \(90.21 \pm 0.19\)& \(70.96 \pm 0.39\)  \\
	GAT & \(91.49 \pm 0.29 \)& \(90.69 \pm 0.10\)& \(80.93 \pm 0.25\)  \\
	SEAL-CI & \(92.93 \pm 0.19 \)& \(92.82 \pm 0.17\)& \(84.74 \pm 0.17\)  \\
	NFP-GCN & \(93.22 \pm 0.09 \)& \(93.07 \pm 0.46\)& \(85.29 \pm 0.38\)  \\
	\textbf{MIRACLE} & {\(\bf{ 98.95 \pm 0.15}\)} & { \(\bf{98.17 \pm 0.06}\)} & { \(\bf{93.20 \pm 0.27}\)} \\
	\bottomrule
\end{tabular}
\label{tabsmall}
\end{table}

\subsubsection{Comparison on the \emph{ChCh-Miner} dataset}
In this part of the experiment, we aim to evaluate our proposed \textbf{MIRACLE} method on datasets with few labeled DDI links. We only compare \textbf{MIRACLE} with graph-based baselines because this dataset lacks similarity-based fingerprints for drug pairs. For the same reason, \textbf{AttGA} does not apply to this dataset. Table \ref{tabmed} shows the results.

Obviously, methods taking multi-view information into consideration like \textbf{SEAL-CI} and \textbf{NFP-GCN} outperform baselines who only use single-view information. However, \textbf{MIRACLE} achieves the best performance and substantially exceeds baselines, demonstrating the superiority of our proposed method on datasets with few labeled data.

The graph contrastive learning component integrates and balances information from different views, \textbf{MIRACLE} learns drug representations better with less labeled DDI links. The proposed method also considers the DDI network's structural information when making the final predicting decisions, which helps the model extracts the most useful information from all dimensional features for DDI prediction. We further verify our points by a small ablation study in which we adjust the training ratio of the dataset in subsection \ref{abla}.

\begin{table}
\centering
\caption{Comparative evaluation results on \emph{ChCh-Miner}}
\begin{tabular}{c|ccc}
	\toprule
	\multirow{2}[0]{*}{\textbf{Algorithm}} & \multicolumn{3}{c}{\textbf{Performance}} \\
\cline{2-4}
		& \textbf{AUROC} & \textbf{AUPRC} & \textbf{F1} \\ \hline
	GCN & \(82.84 \pm 0.61 \)& \(84.27 \pm 0.66\)& \(70.54 \pm 0.87\)  \\
	GIN & \(70.32 \pm 0.87 \)& \(72.41 \pm 0.63\)& \(65.54 \pm 0.97\)  \\
	GAT & \(85.84 \pm 0.23 \)& \(88.14 \pm 0.25\)& \(76.51 \pm 0.38\)  \\
	SEAL-CI & \(90.93 \pm 0.19 \)& \(89.38 \pm 0.39\)& \(84.74 \pm 0.48\)  \\
	NFP-GCN & \(92.12 \pm 0.09 \)& \(93.07 \pm 0.69\)& \(85.41 \pm 0.18\)  \\
	\textbf{MIRACLE} & {\(\bf{ 96.15 \pm 0.29}\)} & { \(\bf{95.57 \pm 0.19}\)} & { \(\bf{92.26 \pm 0.09}\)} \\
	\bottomrule
\end{tabular}
\label{tabmed}
\end{table}

\subsubsection{Comparison on the \emph{DeepDDI} dataset}
To verify our proposed method's scalability, we also conduct experiments on a large-scale dataset, \emph{DeepDDI}, which has plenty of labeled data and DDI information. Table \ref{tablar} compares our proposed approach's performance to the baseline methods. Many baseline approaches utilizing similarity-based fingerprints need plenty of non-structural similarity features, which may be absent in this large scale dataset. Therefore, we only select models who are applicative in this dataset, including \textbf{NN} and \textbf{SSP-MLP}. For the graph-based methods, we neglect \textbf{AttGA} for the lack of many needed drug features. We also ignore experimental results obtained by \textbf{GIN} and \textbf{NFP-GCN} because of the worse performance and the space limitation.

\textbf{MLP-SSP} substantially outperforms \textbf{NN} for the former framework is based on deep neural networks. \textbf{GCN} achieves better performance than \textbf{GAT}, which further demonstrates the inter-view information plays a vital role in DDI predictions. \textbf{SEAL-CI} is second to the proposed method among the baselines proving the superiority of the multi-view graph framework. \textbf{MIRACLE} significantly outperformed other baseline methods in terms of all the three metrics.

\begin{table}
\centering
\caption{Comparative evaluation results on \emph{DeepDDI}}
\begin{tabular}{c|ccc}
	\toprule
	\multirow{2}[0]{*}{\textbf{Algorithm}} & \multicolumn{3}{c}{\textbf{Performance}} \\
\cline{2-4}
		& \textbf{AUROC} & \textbf{AUPRC} & \textbf{F1} \\ \hline
	NN & \(81.81 \pm 0.37 \)& \(80.82 \pm 0.20\)& \(71.37 \pm 0.18\)  \\
	SSP-MLP & \(92.28 \pm 0.18 \)& \(90.27 \pm 0.28\)& \(79.71 \pm 0.16\)  \\
	GCN & \(85.53 \pm 0.17 \)& \(83.27 \pm 0.31\)& \(72.18 \pm 0.22\)  \\
	GAT & \(84.84 \pm 0.23 \)& \(81.14 \pm 0.25\)& \(73.51 \pm 0.38\)  \\
	SEAL-CI & \(92.83 \pm 0.19 \)& \(90.44 \pm 0.39\)& \(80.70 \pm 0.48\)  \\
	\textbf{MIRACLE} & {\(\bf{ 95.51 \pm 0.27}\)} & { \(\bf{92.34 \pm 0.17}\)} & { \(\bf{83.60 \pm 0.33}\)} \\
	\bottomrule
\end{tabular}
\label{tablar}
\end{table}

\begin{figure}[htbp]
\label{fig:ratio}
\centering
\scalebox{0.9}{
\begin{tikzpicture}
\begin{axis}[
    xlabel={Training ratio},
    ylabel={AUROC},
    xmin=0.15, xmax=0.85,
    ymin=0.69, ymax=1.01,
    xtick={0.2,0.4,0.6,0.8},
    ytick={0.7,0.8,0.9,1},
    legend pos=south east,
    ymajorgrids=true,
    grid style=dashed,
    every axis plot/.append style={thick}
]

\addplot[
    color=orange,
    mark=square,
    ]
    coordinates {
    (0.2,0.74)(0.3,0.79)(0.4,0.835)(0.5,0.875)(0.6,0.91)(0.7,0.935)(0.8,0.95)
    };
\addlegendentry{Ens}

\addplot[
    color=purple,
    mark=triangle*,
    ]
    coordinates {
    (0.2,0.785)(0.3,0.821)(0.4,0.860)(0.5,0.881)(0.6,0.897)(0.7,0.91)(0.8,0.92)
    };
\addlegendentry{GCN}

\addplot[
    color=cyan,
    mark=+,
    ]
    coordinates {
    (0.2,0.71)(0.3,0.761)(0.4,0.798)(0.5,0.832)(0.6,0.864)(0.7,0.893)(0.8,0.91)
    };
\addlegendentry{GAT}

\addplot[
    color=teal,
    mark=o,
    ]
    coordinates {
    (0.2,0.833)(0.3,0.856)(0.4,0.872)(0.5,0.887)(0.6,0.903)(0.7,0.918)(0.8,0.93)
    };
\addlegendentry{SEAL-CI}

\addplot[
    color=blue,
    mark=otimes*,
    ]
    coordinates {
    (0.2,0.927)(0.4,0.955)(0.6,0.974)(0.8,0.99)
    };
\addlegendentry{MIRACLE}

\end{axis}
\end{tikzpicture}}
\caption{Results of the compared methods on \emph{ZhangDDI} with different training ratio}
\label{fig:ratio}
\end{figure}
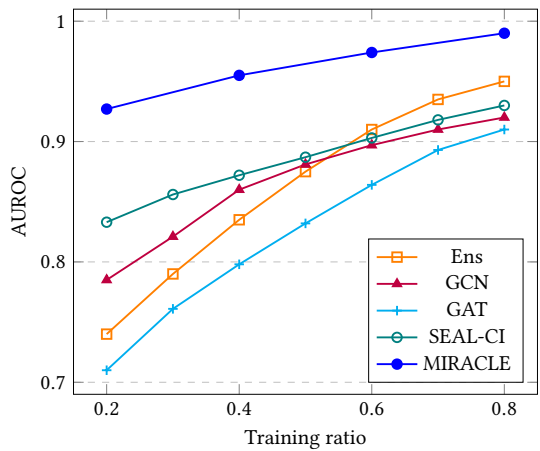

\begin{figure*}[htbp]
\centering
\subfigure[Parameter study of $\alpha$]{
\centering
\scalebox{0.72}{
\begin{tikzpicture}

\begin{semilogxaxis}[
    xlabel={\Large{$\alpha$}},
    xmode=log,
    xmin=0.005, xmax=2000,
    ymin=0.6, ymax=1.1,
    xtick={0.01,0.1,1,10,100,1000},
    ytick={0.7,0.8,0.9,1},
    log ticks with fixed point,
    x tick label style={/pgf/number format/1000 sep=\,},
    legend pos=south east,
    ymajorgrids=true,
    grid style=dashed,
]

\addplot[
    color=cyan,
    mark=square*,
    ]
    coordinates {
    (0.01,0.974)(0.1,0.984)(1,0.987)(10,0.988)(100,0.990)(1000,0.975)
    };
\addlegendentry{AUROC}

\addplot[
    color=magenta,
    mark=otimes* ,
    ]
    coordinates {
    (0.01,0.924)(0.1,0.951)(1,0.975)(10,0.979)(100,0.983)(1000,0.965)
    };
\addlegendentry{AUPRC}

\addplot[
    color=blue,
    mark=triangle*,
    ]
    coordinates {
    (0.01,0.821)(0.1,0.852)(1,0.895)(10,0.924)(100,0.932)(1000,0.915)
    };
\addlegendentry{F1}

\end{semilogxaxis}
\end{tikzpicture}}
\label{fig:a}
}
\subfigure[Parameter study of $\beta$]{
\centering
\scalebox{0.72}{
\begin{tikzpicture}

\begin{axis}[
    title={ },
    xlabel={\Large{$\beta$}},
    xmin=0.14, xmax=1.46,
    ymin=0.6, ymax=1.1,
    xtick={0.2,0.4,0.6,0.8,1,1.2,1.4},
    ytick={0.7,0.8,0.9,1},
    legend pos=south east,
    ymajorgrids=true,
    grid style=dashed,
]

\addplot[
    color=cyan,
    mark=square*,
    ]
    coordinates {
    (0.2,0.954)(0.4,0.966)(0.6,0.978)(0.8,0.990)(1.0,0.987)(1.2,0.975)(1.4,0.963)
    };
\addlegendentry{AUROC}

\addplot[
    color=magenta,
    mark=otimes* ,
    ]
    coordinates {
    (0.2,0.911)(0.4,0.934)(0.6,0.952)(0.8,0.987)(1.0,0.987)(1.2,0.966)(1.4,0.958)
    };
\addlegendentry{AUPRC}

\addplot[
    color=blue,
    mark=triangle*,
    ]
    coordinates {
    (0.2,0.863)(0.4,0.882)(0.6,0.898)(0.8,0.932)(1.0,0.923)(1.2,0.889)(1.4,0.867)
    };
\addlegendentry{F1}

\end{axis}
\end{tikzpicture}}
\label{fig:b}
}
\subfigure[Parameter study of $d_g$]{
\centering
\scalebox{0.72}{
\begin{tikzpicture}

\begin{semilogxaxis}[
    title={ },
    xlabel={\Large{$d_g$}},
    xmode=log,
    xmin=1.5, xmax=700,
    ymin=0.6, ymax=1.1,
    xtick={2,4,8,16,32,64,128,256,512},
    ytick={0.7,0.8,0.9,1},
    log ticks with fixed point,
    x tick label style={/pgf/number format/1000 sep=\,},
    legend pos=south east,
    ymajorgrids=true,
    grid style=dashed,
    log basis x={2}
]

\addplot[
    color=cyan,
    mark=square*,
    ]
    coordinates {
    (2,0.887)(4,0.892)(8,0.897)(16,0.915)(32,0.933)(64,0.949)(128,0.974)(256,0.990)(512,0.978)
    };
\addlegendentry{AUROC}

\addplot[
    color=magenta,
    mark=otimes* ,
    ]
    coordinates {
    (2,0.827)(4,0.832)(8,0.847)(16,0.875)(32,0.903)(64,0.923)(128,0.964)(256,0.982)(512,0.968)
    };
\addlegendentry{AUPRC}

\addplot[
    color=blue,
    mark=triangle*,
    ]
    coordinates {
    (2,0.747)(4,0.772)(8,0.807)(16,0.835)(32,0.843)(64,0.879)(128,0.924)(256,0.932)(512,0.918)
    };
\addlegendentry{F1}

\end{semilogxaxis}
\end{tikzpicture}}
\label{fig:c}
}
\caption{Parameter sensitivity study on \emph{ZhangDDI}}
\label{fig:para}
\end{figure*}
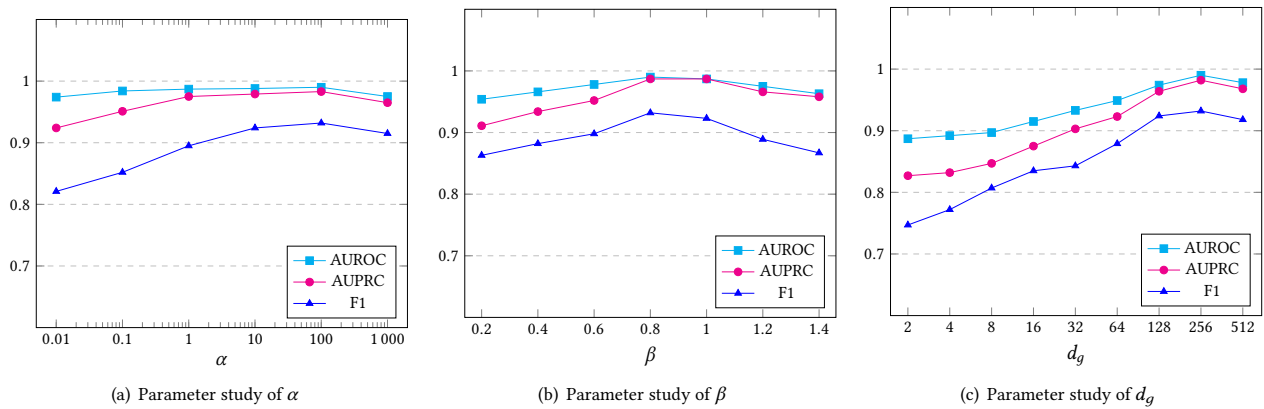

\subsection{Ablation Study}\label{abla}
\begin{figure}[htbp]
\centering
\subfigure[-C]{
\includegraphics[width=1.62in]{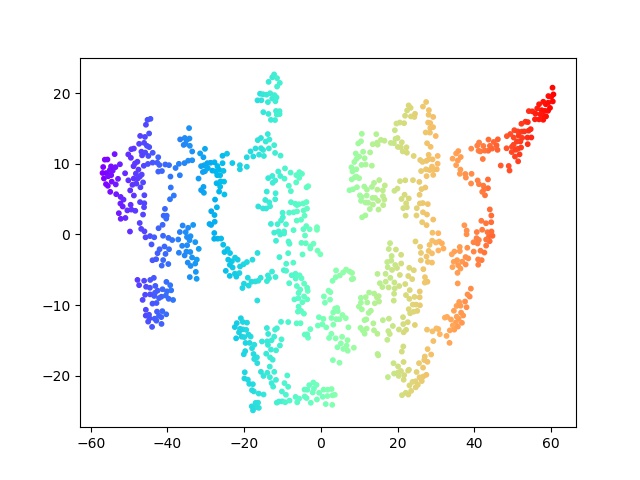}}
\subfigure[MIRACLE]{
\includegraphics[width=1.62in]{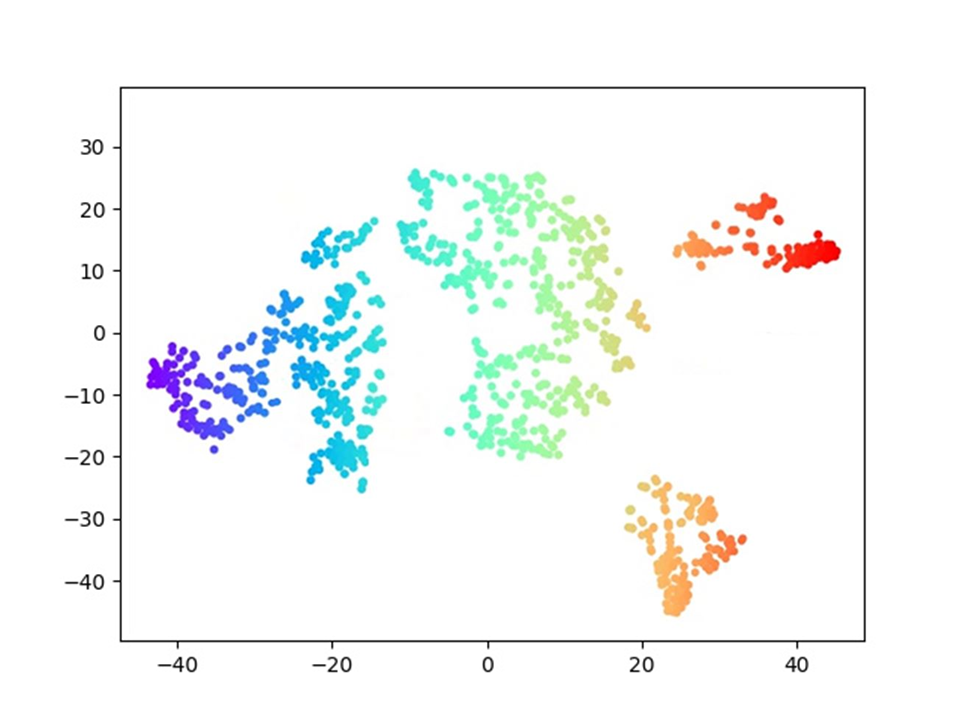}}
\caption{Visualization of drug embeddings on \emph{ChCh-Miner}. (The drug embeddings visualized in the left picture is learned by MIRACLE without the contrastive learning component, and those in the right one is learned directly by MIRACLE.)}
\label{fig:tsne}
\end{figure}

We conduct ablation experiments on the \emph{ZhangDDI} and the \emph{ChCh-Miner} dataset to validate the contrastive learning component's effectiveness in our model. The experimental results are reported in Table \ref{tab:abla}. To better understand the differences between MIRACLE with and without the contrastive learning component, we visualize their drug embeddings using the visualization tool t-SNE\cite{maaten2008visualizing}, which projects each method's learned embeddings in a 2D space. Figure \ref{fig:tsne} shows the results on the \emph{ChCh-Miner} dataset. From the figure, we can observe that MIRACLE effectively learns drug embeddings with the contrastive learning component. We also analyze our proposed method's robustness with a different number of labeled training data by adjusting the training set ratio on the \emph{ZhangDDI} dataset. As seen in Figure \ref{fig:ratio}, MIRACLE with the contrastive learning component can also achieve high performance, validating its superiority in the scenario with few labeled data.

\begin{table}
\centering
\caption{Ablation experimental results}
\scalebox{0.74}{
\begin{tabular}{ccccc}
	\toprule
	\multirow{2}*{\textbf{\Large{Dataset}}} & \multirow{2}*{\textbf{\Large{Algorithm}}} & \multicolumn{3}{c}{\textbf{\Large{Performance}}} \\
\cline{3-5}
		& & \textbf{AUROC} & \textbf{AUPRC} & \textbf{F1} \\ \hline
	\specialrule{0em}{2pt}{0pt}
	\multirow{2}*{\Large{ZhangDDI}} & \Large{-C} & \Large{\(96.62 \pm 0.17\)} & \Large{\(92.15 \pm 0.31\)} & \Large{\(89.42 \pm 0.22 \)}  \\
	& \Large{MIRACLE} & \Large{\(98.95 \pm 0.15\)} & \Large{\(98.17 \pm 0.06\)} & \Large{\(93.20 \pm 0.27 \)}  \\ \hline
	\specialrule{0em}{2pt}{0pt}
	\multirow{2}*{\Large{ChCh-Miner}} & \Large{-C} & \Large{\(93.37 \pm 0.40\)} & \Large{\(94.84 \pm 0.38\)} & \Large{\(90.38 \pm 0.15 \)}  \\
	& \Large{MIRACLE} & \Large{\(96.15 \pm 0.29\)} & \Large{\(95.57 \pm 0.19\)} & \Large{\(92.26 \pm 0.09 \)}  \\
	\bottomrule
\end{tabular}}
\label{tab:abla}
\end{table}

\subsection{Parameter Sensitivity}
In our model in equation \ref{loss}, there are two major parameters $\alpha$ and $\beta$. In this subsection, we evaluate the impacts of them, together with the dimensionality of drug embeddings $d_g$ on the \emph{ZhangDDI} dataset. Figure \ref{fig:a} and \ref{fig:b} show the results by changing one parameter while fixing another one.

First, we vary $\alpha$ by \{\emph{0.01, 0.1, 1, 10, 100, 1000}\}, and fix $\beta = 0.8$, $d_g = 256$. Here, for parameter study purpose, $\beta$ is set to its optimal value on this dataset instead of the default value 1. From the figure, our method is quite stable in a wide range of $\alpha$ and achieves the best performance when $\alpha = 100$ in terms of \emph{AUROC}. Next, we vary $\beta$ by \{\emph{0.2, 0.4, 0.6, 0.8, 1, 1.2, 1.4}\} with $\alpha = 100$ and $d_g = 256$. As can be seen, the near optimal performance at $\beta = 0.8$ justifies our parameter setting. Overall, $\alpha$ and $\beta$ are stable w.r.t these parameters. Moreover, the non-zero choices of $\alpha$ and $\beta$ demonstrate the importance of the loss terms in our model.

To evaluate $d_g$, we vary it from 2 to 512, and fix $\alpha = 10$ and $\beta = 0.8$. The results are shown in Figure \ref{fig:c}. From the figure, MIRACLE is robust to $d_g$. Specifically, when $d_g$ is small, the \emph{AUROC} increases because higher dimensionality can encode more useful information. When $d_g$ reaches its optimal value, the accuracy begins to drop slightly because a too high dimensionality may introduce redundant and noisy information that can harm the classification performance.

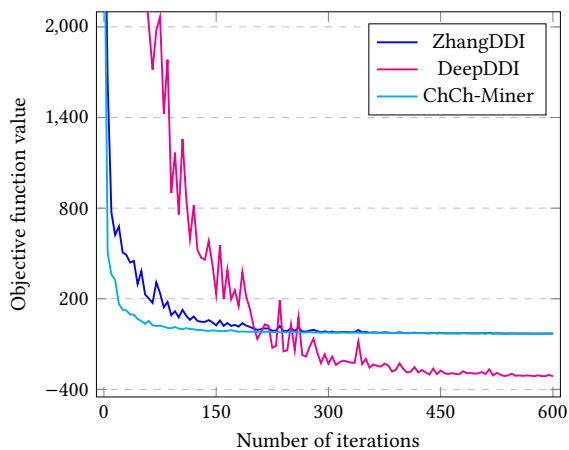
\begin{figure}[htbp]
\label{fig:conv}
\centering
\scalebox{0.9}{
\begin{tikzpicture}
\begin{axis}[
    xlabel={Number of iterations},
    ylabel={Objective function value},
    xmin=-10, xmax=610,
    ymin=-450, ymax=2100,
    xtick={0,150,300,450,600},
    ytick={-400,200,800,1400,2000},
    legend pos=north east,
    ymajorgrids=true,
    grid style=dashed,
    no markers,
    every axis plot/.append style={thick}
]

\addplot[
    color=blue,
    ]
    coordinates {
    (0,3339.9)(5,1637)(10,771.1)(15,622.3)(20,678.7)(25,507.0)(30,489.9)(35,440.0)(40,451.0)(45,296.7)(50,382.6)(55,228.6)(60,202.0)(65,172.4)(70,308.4)(75,237.4)(80,143.3)(85,179.2)(90,90.79)(95,117.5)(100,76.62)(105,126.5)(110,87.33)(115,60.29)(120,82.59)(125,52.99)(130,47.87)(135,46.66)(140,58.89)(145,43.40)(150,23.77)(155,56.17)(160,20.59)(165,39.92)(170,20.40)(175,26.43)(180,15.98)(185,38.45)(190,22.44)(195,13.99)(200,2.526)(205,-5.84)(210,-1.22)(215,3.772)(220,2.873)(225,-11.7)(230,-10.4)(235,19.86)(240,-14.1)(245,-13.4)(250,3.807)(255,-13.4)(260,8.345)(265,-16.6)(270,-17.6)(275,-10.7)(280,-5.66)(285,-14.9)(290,-21.8)(295,-15.9)(300,-22.4)(305,-17.6)(310,-23.0)(315,-21.1)(320,-20.2)(325,-20.8)(330,-21.6)(335,-21.6)(340,-7.30)(345,-22.4)(350,-19.1)(355,-24.9)(360,-22.7)(365,-23.7)(370,-24.2)(375,-22.1)(380,-23.4)(385,-27.5)(390,-26.3)(395,-22.0)(400,-25.7)(405,-28.0)(410,-27.5)(415,-24.5)(420,-28.0)(425,-26.5)(430,-28.0)(435,-29.4)(440,-25.5)(445,-29.1)(450,-26.7)(455,-28.5)(460,-29.1)(465,-28.4)(470,-29.0)(475,-29.2)(480,-27.2)(485,-28.6)(490,-28.4)(495,-28.8)(500,-28.6)(505,-27.8)(510,-29.1)(515,-26.0)(520,-30.0)(525,-28.3)(530,-29.6)(535,-30.4)(540,-30.0)(545,-28.8)(550,-30.2)(555,-30.0)(560,-30.3)(565,-30.0)(570,-30.1)(575,-30.5)(580,-29.9)(585,-30.4)(590,-30.5)(595,-29.6)(600,-30.6)
    };
\addlegendentry{ZhangDDI}

\addplot[
    color=magenta,
    ]
    coordinates {
    (5,10838.8)(10,7700.9)(15,6213.2)(20,6777.8)(25,5060.4)(30,4888.9)(35,4391.1)(40,4500.7)(45,2957.9)(50,3817.8)(55,2278.1)(60,2011.5)(65,1715.9)(70,1976.4)(75,2066.7)(80,1425.6)(85,1783.2)(90,899.89)(95,1167.4)(100,758.39)(105,1257.1)(110,865.96)(115,595.60)(120,818.38)(125,522.52)(130,471.63)(135,459.25)(140,581.61)(145,426.29)(150,230.36)(155,554.22)(160,198.66)(165,392.36)(170,196.52)(175,257.14)(180,152.55)(185,377.16)(190,217.27)(195,132.82)(200,17.990)(205,-65.64)(210,-19.41)(215,30.535)(220,21.626)(225,-124.6)(230,-111.4)(235,190.84)(240,-148.7)(245,-141.1)(250,30.870)(255,-141.9)(260,76.987)(265,-172.8)(270,-182.8)(275,-115.6)(280,-64.64)(285,-154.2)(290,-226.5)(295,-168.3)(300,-229.3)(305,-182.8)(310,-238.1)(315,-219.8)(320,-209.4)(325,-216.3)(330,-222.7)(335,-224.3)(340,-84.06)(345,-232.5)(350,-198.3)(355,-256.7)(360,-235.2)(365,-244.6)(370,-249.2)(375,-229.2)(380,-241.7)(385,-282.0)(390,-270.9)(395,-227.5)(400,-264.4)(405,-287.6)(410,-282.3)(415,-252.7)(420,-287.7)(425,-271.9)(430,-287.7)(435,-300.9)(440,-262.4)(445,-299.0)(450,-275.3)(455,-292.5)(460,-298.3)(465,-292.2)(470,-298.0)(475,-299.5)(480,-279.6)(485,-293.0)(490,-291.4)(495,-295.6)(500,-293.6)(505,-285.1)(510,-298.8)(515,-267.4)(520,-307.6)(525,-290.3)(530,-302.7)(535,-312.3)(540,-307.2)(545,-296.1)(550,-309.8)(555,-307.5)(560,-310.7)(565,-307.5)(570,-309.2)(575,-312.7)(580,-306.8)(585,-312.2)(590,-311.9)(595,-304.0)(600,-313.3)
    };
\addlegendentry{DeepDDI}

\addplot[
    color=cyan,
    ]
    coordinates {
    (0,3110)(5,502.984)(10,360.99)(15,323.94)(20,165.45)(25,127.49)(30,122.51)(35,96.519)(40,95.206)(45,66.127)(50,52.660)(55,34.214)(60,53.819)(65,23.524)(70,21.033)(75,20.072)(80,15.955)(85,6.7259)(90,4.9049)(95,12.953)(100, 2.723)(105,-1.458)(110,5.6418)(115,-0.310)(120,-2.105)(125,-6.699)(130,-7.134)(135,-10.47)(140,-13.09)(145,-9.325)(150,-11.05)(155,-13.94)(160,-14.78)(165,-12.28)(170,-7.429)(175,-13.55)(180,-18.85)(185,-16.32)(190,-17.88)(195,-18.50)(200,-18.14)(205,-19.26)(210,-20.71)(215,-19.75)(220,-19.22)(225,-21.08)(230,-20.99)(235,-21.17)(240,-21.28)(245,-22.07)(250,-20.05)(255,-20.87)(260,-22.85)(265,-23.98)(270,-22.60)(275,-21.30)(280,-23.57)(285,-23.25)(290,-24.62)(295,-23.17)(300,-24.19)(305,-24.57)(310,-25.73)(315,-24.35)(320,-24.04)(325,-25.54)(330,-24.33)(335,-23.84)(340,-24.12)(345,-25.79)(350,-26.25)(355,-26.92)(360,-26.45)(365,-25.87)(370,-26.99)(375,-25.82)(380,-26.69)(385,-25.66)(390,-26.71)(395,-27.46)(400,-27.01)(405,-27.90)(410,-27.14)(415,-27.24)(420,-27.43)(425,-27.66)(430,-25.56)(435,-26.84)(440,-27.24)(445,-28.02)(450,-28.44)(455,-28.45)(460,-28.61)(465,-28.84)(470,-28.89)(475,-28.88)(480,-27.10)(485,-28.92)(490,-29.43)(495,-29.18)(500,-29.40)(505,-28.56)(510,-30.08)(515,-28.80)(520,-29.63)(525,-29.90)(530,-29.87)(535,-30.42)(540,-30.42)(545,-30.62)(550,-30.44)(555,-30.34)(560,-30.80)(565,-30.62)(570,-30.54)(575,-31.27)(580,-31.15)(585,-31.22)(590,-30.25)(595,-31.90)(600,-29.88)
    };
\addlegendentry{ChCh-Miner}

\end{axis}
\end{tikzpicture}}
\caption{Convergence evaluation}
\label{fig:conv}
\end{figure}

\subsection{Convergence Evaluation}
In this subsection, we study the proposed MIRACLE algorithm's performance in terms of the number of iterations before converging to a local optimum. Figure \ref{fig:conv} shows the value of the objective function $\mathscr{L}$ in equation (\ref{loss}) with respect to the number of iterations on different datasets. From the figure, we observe the objective function value decreases steadily after many iterations.

\section{Related work}
\textbf{DDI prediction.} There is little doubt that the most accurate way for predicting DDIs is \textit{in vivo} or \textit{in vitro} trials. But such bioassay is practicable only on a small set of drugs and limited by the experimental environment\cite{duke2012literature}. With the accumulation of biomedical data, many machine learning methods are proposed for DDI prediction. To the best of our knowledge, most existing DDI prediction methods could be divided into similarity-based and graph-based.

The similarity-based methods adopt the assumption: \textbf{drugs sharing similar chemical structures are prone to share similar DDI.} \citet{vilar2012drug, vilar2014similarity} generate DDI candidates by comparing the Tanimoto coefficient between human-curated fingerprint vectors of drugs. \citet{celebi2015prediction} employs a rooted page rank algorithm that integrates therapeutic, genomic, phenotypic, and chemical similarity to discover novel DDIs. \citet{zhang2015label} proposes a label propagation framework considering immediate structure similarity from PubChem and high-order similarity from divergent clinical databases. \citet{huang2013systematic, luo2014ddi} measure the connections between drugs and targets to predict pharmacodynamic DDIs, which could be regarded as another aspect of drug similarity. \citet{ryu2018deep} proposes DeepDDI, which applies a feed-forward neural network to encode structural similarity profile (SSP) of drugs to predict DDI. Although similarity-based methods yield robust results on DDI prediction tasks, these methods may have limitations on digging novel DDIs, i.e., not scalable. Starting with known drug interactions, it is far from sufficient to use similarity criteria to model complex DDIs, let alone the number of DDIs identified is still sparse. For high-hanging fruits, more prior knowledge and higher-level representations are required for novel DDI detection.

Recently, graph-based methods like neural fingerprint\cite{duvenaud2015convolutional}, message passing neural networks\cite{DBLP:conf/icml/GilmerSRVD17} and graph representation learning-based framework\cite{hamilton2017inductive} have proved to be successful on molecular tasks, several graph neural networks (GNNs) are proved to have a good DDI prediction power. \citet{zhang2015label, zhang2017predicting} make DDI predictions based on nearest neighbors, though these works learn representations from multiple prior information, but failed to take relations of DDIs into account, such as transitivity of similarities. \citet{10.5555/3304222.3304251} uses attentive graph auto-encoders\cite{kipf2016variational} to integrates heterogeneous information from divergent drug-related data sources. Decagon\cite{zitnik2018modeling} introduces another graph auto-encoder, making an end-to-end generation of link prediction predictions in the drug-protein network. This work shows that deep GNNs outperform traditional shallow architectures significantly, such as static fingerprints\cite{duvenaud2015convolutional}, DeepDDI\cite{ryu2018deep}, and traditional graph embedding methods\cite{perozzi2014deepwalk, zong2017deep}. More recently, \citet{lin2020kgnn} merges several datasets into a vast knowledge graph(KG) with 1.2 billion triples. Through KGNN layers which embed 2-hop local structures of drugs, this model outperforms a previous KG-based method KG-ddi\cite{karim2019drug} significantly. With the help of such graph-based models, higher-level representations are learned from a variety of graph-structured data and improve the performance on the DDI prediction task.

\textbf{Neural network for graphs.} Graph neural networks extend the convolutional neural networks to non-Euclidean spaces, providing a more natural and effective way for modeling graph-structured data\cite{DBLP:conf/icml/GilmerSRVD17}. In recent years, numerous powerful GNN architectures are flourished\cite{scarselli2008graph, li2015gated,duvenaud2015convolutional,kearnes2016molecular,kipf2016semi,battaglia2016interaction,hamilton2017inductive,velivckovic2017graph, xu2018representation,zhang2018end,xu2018powerful}. With the development of various GNNs and their variants, modern graph representation learning-based methods have achieved state-of-the-art performances on link prediction, node classification, graph classification. Generally, a typical GNN follows a three-step message passing scheme to generate node or graph level representation. 1) Aggregate step: integrating information from neighboring nodes which serve as messages; 2) Combine step: updating the representation of nodes based on collected messages in 1); 3) Readout step: applying a permutation invariant function that produces graph-level representation from node-level representations. In each layer of GNN, node features are usually updated by its immediate neighbors. Thus, after $k$ iterations (layers), a node's embedding vector encodes the $k$-hop subgraph's topological information. Finally, a graph-level embedding reflecting overall structure could be computed by readout function if needed.

Recently, contrastive learning revitalizes in many fields, and graph representation learning is no exception. Inspired by previous success in visual representation learning, \citet{velivckovic2018deep} proposes a MI maximization objective to maximize the MI between node embeddings and the graph embedding by discriminating nodes in the original graph from nodes in a corrupted graph. \citet{sun2019infograph} extends the MI maximization objective to better learn the graph-level representation by maximizing the MI between the graph-level representation and the representations of substructures of different scales. \citet{zhu2020deep} proposes a new method to generate different graph views by removing edges and masking features and maximizing the agreement of node embeddings between two graph views. Inspired by the information maximization principle, we apply the MI maximization objective in balancing information from different views in drug embeddings.

\section{conclusion}
To better integrate the rich multi-view graph information in the DDI network, we propose MIRACLE for the DDI prediction task in this paper. MIRACLE learns drug embeddings from a multi-view graph perspective by designing an end-to-end framework that consists of a bond-aware message passing network and a GCN encoder. Then, a novel contrastive learning-based strategy has been proposed to balance information from different views. Also, we design two predictors from both views to fully exploit the available information. Through extensive experiments on various real-life datasets, we have demonstrated that the proposed MIRACLE is both effective and efficient.